 \documentclass[pmlr,twocolumn,10pt]{jmlr} 

 \usepackage{longtable}

\usepackage{booktabs}
 \usepackage{pgfplots}
\usepackage{siunitx}
\usepackage[acronym]{glossaries}

\theorembodyfont{\upshape}
\theoremheaderfont{\scshape}
\theorempostheader{:}
\theoremsep{\newline}

\jmlrvolume{LEAVE UNSET}
\jmlryear{2023}
\jmlrsubmitted{LEAVE UNSET}
\jmlrpublished{LEAVE UNSET}
\jmlrworkshop{Machine Learning for Health (ML4H) 2023} 

 \title[Language Models for Epic EHR Audit Logs]{Autoregressive Language Models For Estimating the Entropy of Epic EHR Audit Logs}

\author{%
\Name{Benjamin C. Warner} \Email{b.c.warner@wustl.edu}\\ 
\Name{Thomas Kannampallil} \Email{thomas.k@wustl.edu}\\
\Name{Seunghwan Kim} \Email{seunghwan.kim@wustl.edu}\\
\addr{Washington University in St. Louis}
}

\glsdisablehyper

\begin{document}

\newacronym{al}{audit log}{EHR audit log}
\newacronym{lm}{LM}{language model}
\newacronym{llm}{LLM}{large language model}
\newacronym{tlm}{tabular LM}{tabular language model}
\newacronym{icu}{ICU}{intensive care unit}
\newacronym{ehr}{EHR}{electronic health record}
\newacronym{phi}{PHI}{protected health information}

\newglossaryentry{mn}
{
    name=\texttt{METRIC\_NAME},
    description={Action performed}
}

\newglossaryentry{at}
{
    name=\texttt{ACCESS\_TIME},
    description={Time accessed}
}

\newglossaryentry{pid}
{
    name=\texttt{PAT\_ID},
    description={Action performed}
}

\newglossaryentry{fv}
{
    name={field vocab},
    description={Vocabulary for a field}
}

\newglossaryentry{gv}
{
    name={global vocab},
    description={Vocabulary for tabular data}
}

\newglossaryentry{ce}{
    name={cross-entropy},
    description={Cross entropy}
}

\newglossaryentry{td}{
    name={time-delta},
    description={time-delta}
}

\maketitle

\begin{abstract}
\Acrlongpl{al} are a highly granular stream of events that capture clinician activities, and is a significant area of interest for research in characterizing clinician workflow on the \gls{ehr}. Existing techniques to measure the complexity of workflow through \glspl{al} involve time- or frequency-based cross-sectional aggregations that are unable to capture the full complexity of a \gls{ehr} session. We briefly evaluate the usage of transformer-based \gls{tlm} in measuring the entropy or disorderedness of action sequences within workflow and release the evaluated models publicly.
\end{abstract}


\glsdisablehyper

\section{Introduction} \label{sec:introduction}
Modern day clinical work  relies heavily on documentation on \glspl{ehr} \citep{jha2010ehr}. Raw \acrlongpl{al}\glsunset{al} are trails of clinician activity on the \gls{ehr} and capture a fine-grained representation of electronic interaction within the clinical workflow. However, the high dimensionality and lack of context in \acrlongpl{al} makes it challenging to use it to effectively  evaluate the complexity of clinical workflow and work activities. Current methods for quantifying clinical workflow complexities using \glspl{al} tend to be limited to time-based or frequency-based measure of action events \citep{kannampallil2023using}, which does not take into account for the underlying temporal relationships between \gls{ehr} actions and the sequential nature of \gls{ehr}-based interactions.

Being able to  quantify the disorderedness in \gls{al} action sequences could be useful in accurately characterizing the complexity of clinical workflow. To the best of our knowledge, there is no standardized technique currently available for doing so. To measure the disorder or entropy of actions captured in \acrlongpl{al} we trained several \glspl{tlm} and use them to estimate the cross-entropy of each action event as a proxy of action entropy in an \gls{al} action sequence. 

\subsection{Contributions}

Our contributions are the following:

\begin{itemize}
    \item Publicly-available autoregressive \acrlongpl{lm} for evaluating the entropy of user interaction sequences using \acrlongpl{al}.
    \item A comparison and discussion of different \acrlongpl{lm} architectures for assessing \glspl{al}.
    \item An evaluation of the \gls{ce} of an \gls{al} sequence as a proxy for measuring workflow complexity.
\end{itemize}

We make our code available on GitHub here: \url{https://github.com/bcwarner/audit-log-lm}.

Model weights are available on Hugging Face and reachable through our GitHub.

\section{Related Work}

\subsection{\Acrlongpl{al}}


As mandated by the Health Insurance Portability and Accountability Act (HIPAA), all EHR-based activities are recorded to monitor access to protected patient health information. Raw \acrlongpl{al} are user-centric time-series stream of action events resulting from  click-level user interaction on the \gls{ehr}, and captures a broad range of actions and their data \citep{adler2020ehr}. Each row of \glspl{al} represents an action event. Such unique characteristics of \acrlongpl{al} allows for studying clinical workflow processes at scale. Currently, there are no standard metrics for assessing workflow and work activities from \glspl{al} and considerable effort is often expended on designing, computing, and validating a myriad of different measures \citep{kannampallil2023using}. However, most currently proposed metrics regarding \gls{ehr} use based on cross-sectional aggregation of individual action events within a sequence and limited work has been done to incorporate inter-event transitions and relationships to more accurately quantify the workflow complexity.

Previous \gls{al} analyses that attempted to incorporate such sequential characteristics of \gls{ehr}-based interactions have utilized \glspl{lm}. Such studies leverage training word2vec \citep{mikolov2013efficient} on action events, and then using the embeddings for various downstream tasks such as clustering \citep{lou2023characterizing, jones2020learning} or in supervised learning of physician states such as burnout \citep{liu2022hipal}.

\subsection{Entropy}

\citet{shannon1948mathematical} introduced the measure of entropy for measuring the number of bits required to represent a given state. The measure of entropy is dependent on the probability function $P$, whose true behavior is unknown. For this reason, we generally evaluate cross-entropy, which evaluates a model $\hat{P}$ of the probability function. Cross-entropy has the useful property that it is an upper-bound for true entropy, and decreases monotonically as $\hat{P}$ improves. Entropy has generally been used to represent the amount of \textit{surprise} or \textit{disorder}, as lower probability states will have higher entropy. In natural language processing tasks, cross-entropy is often reported through \textit{perplexity}, which is the exponentiation of cross-entropy, as it represents the geometric average of the number of possible outcomes of $\hat{P}$ \citep{jurafskyspeech}.

Using raw \gls{ehr} \gls{al} data, cross-entropy calculated for each \gls{ehr}-based action can be used as proxy of representing complexity of clinical workflow. We expect higher entropy states to have higher action sequence complexity, as action sequences that deviate from commonly-used sequences represent greater disorder within the workflow. Conversely, more common action sequences will have lower complexity and hence, less entropy.

\subsection{\Acrlong{lm}}

\Acrfullpl{lm} represent a model $P$ of words or characters in a given language, or at a more fundamental level, a model of a sequence. Language models can be divided into two broad classes: autoregressive and bidirectional. In the autoregressive case, \glspl{lm} attempt to learn $P(x_i|x_{i - k}...x_{i - 1})$, the next word in a sequence given a context of up to $k$ words \citep{radford2018improving}, and use cross-entropy as a training objective. A bidirectional model attempts to predict words from both preceding and succeeding tokens \citep{devlin2018bert}. We focus primarily on autoregresssive language models here since our principal interest is to use these models for calculating cross-entropy directly. 

Most recent \glspl{lm} are based upon the transformer architecture introduced in \citet{vaswani2017attention}, and many of these variations attempt to address different architectural issues. We will examine three different architectures: GPT-2 \citep{radford2018improving}, RWKV \citep{peng2023rwkv}, and LLaMA \citep{touvron2023llama}. We choose to evaluate GPT-2 for its generality and simplicity, RWKV for its linearized attention mechanism, and LLaMA for its recency in architectural advancements.

\subsection{\Acrlongpl{tlm}}

\Acrlongpl{tlm} are \glspl{lm} that are trained to predict tabular time-series data. This is done by converting each entry $x_{i, j}$ of a table into a sequence of tokens $t_{1, 1},...,t_{1, n},t_{2, 1},...,t_{m, n}$, and then training or inferring from the tokenized data. Each field in a tabular dataset has its own \textit{\gls{fv}}, which can be concatenated together to form a \textit{\gls{gv}}. To evaluate the loss for a \gls{tlm}, we take the \gls{ce} over the logits for each \gls{fv}, and then reduce all field losses together \citep{padhi2021tabular}.

This vocabulary scheme mandates that all fields have a finite set of tokens from which to choose from, and requires that fields that are not finite, namely continuous fields, to be processed. The approach \citet{padhi2021tabular} propose is to quantize the values, where bins are computed for a given field and continuous values are then mapped to these bins. An alternative approach that \citet{solatorio2023realtabformer} propose is to serialize and partition the formatted string value into finite tokens, which may require multiple columns. 

\section{Methodology} \label{sec:methods}

\subsection{Preprocessing}

The dataset consisted of retrospectively collected all raw \gls{al} events of clinicians in 2019, for those who worked across four surgical \glspl{icu} at least once at 
Barnes-Jewish Hospital, a large academic medical center in St. Louis, MO, USA. 
This study was approved by the institutional review board of Washington University (IRB\# 202009032) with a waiver of informed consent.

\Glspl{al} contain many features, and we restrict the \glspl{lm} discussed to predict only three fields: \gls{mn}, \gls{at}, \gls{pid}. \Gls{mn} is one of $4,037$ English-level descriptions of the action performed (e.g. ``\texttt{Radiology Front Desk Orders Report accessed}''). \gls{pid} is a unique identifier for the patient that was interacted with at a given action if present. \gls{at} represents the date and time of an event with second-level precision.

To prepare the \gls{al} data for tokenization, several preprocessing steps are performed. There are instances where a single EHR-based action triggers multiple action events to be recorded in the \gls{al} with an identical \gls{at} timestamp. To account for those, we further sort by the \texttt{ACCESS\_INSTANT} field, which has sub-second precision and lower super-second accuracy. Upon sorting, we convert the \gls{at} feature to \glspl{td}. \Glspl{al} are then subdivided into demarcated shifts of work, with \gls{at} zeroed after a predetermined gap. We use a gap of 6 hours of inactivity to split \glspl{al} into shifts. 

We then convert \gls{pid} into identifiers indicating their appearance in a shift. To ensure that the field vocab for the \gls{pid} is finite, we limit the patient count to 128 per shift.

After splitting the data into shifts, we then split the \gls{al} data into \textit{sessions}, which \citet{ouyang2016} define as periods of activity with gaps no greater than 5 minutes. Each of these sessions represents an \gls{ehr}-based interaction sequence of related clinician work behaviors in a clinical setting.

\Glspl{td} are then quantized into a series of logarithmically-spaced bins of 5 intervals ranging from 0 to 240 seconds. We use logarithmic spacing as this will capture the difference between shorter and longer actions, while minimizing the number of bins the \gls{lm} must learn. 
After this, we tokenize each field. Sessions longer than the sequence length supported by the model are divided into chunks of the appropriate length.

\subsection{Training}

Our training strategy is to randomly stratify by individual clinicians and then shuffle examples in each set, with 114 clinicians appearing in the training set and 24 clinicians appearing in both the test and validation set. With the aforementioned preprocessing strategy, there are 398,412 training examples. We train with a batch size of 2 and use 4 gradient accumulation steps for 5 epochs on all models, and where available we use TensorFloat-32 operations. We optimize the \glspl{lm} discussed here using Sophia, which has been shown experimentally to be significantly faster than Adam \citep{liu2023sophia}. 

\begin{figure*}
    \centering
    \includegraphics[width=0.8\textwidth]{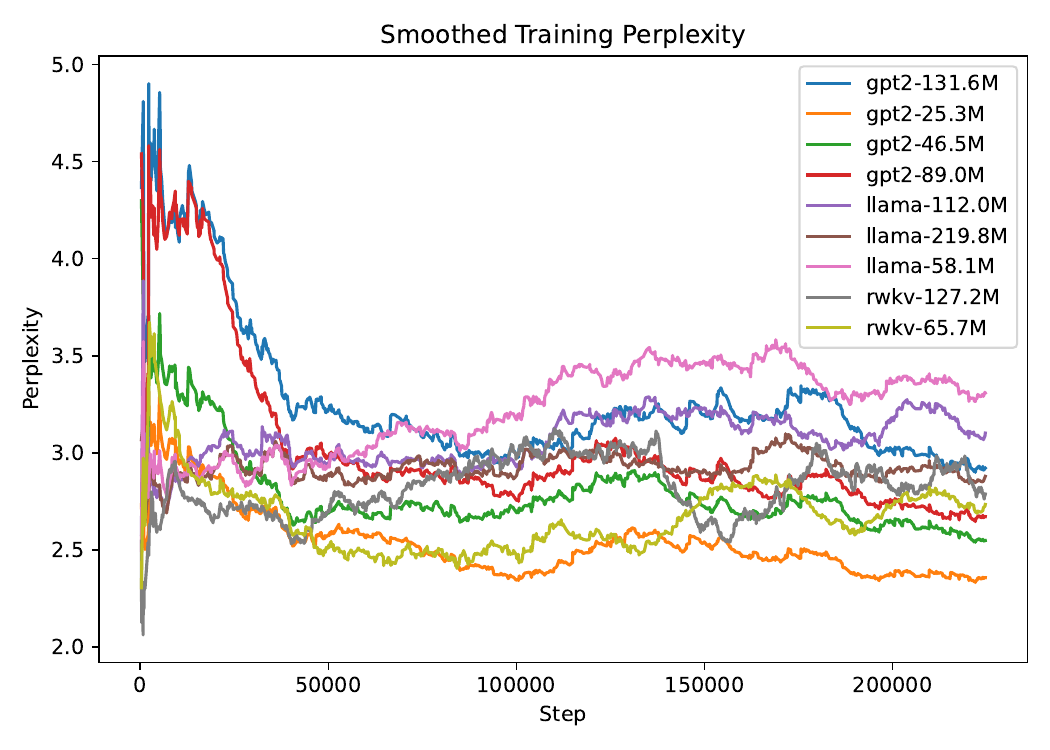}
    \caption{Training loss, with exponentially weighted-smoothing ($\alpha = 0.01$).}
    \label{fig:train_loss}
\end{figure*}

The training loss for our tested models (with exponentially weighted-smoothing for visibility) can be seen above in \figureref{fig:train_loss}. We train for 5 epochs, although it appears that 1-2 epochs may be enough to reach convergence.





%
%

\section{Results} \label{sec:results}

\subsection{Audit Log Generation Performance}

As there is no literature evaluating generative model performance for \glspl{al}, we suggest the usage of two categories of metrics: next-action prediction accuracy and ROUGE scores \citep{lin-2004-rouge}. We evaluate both of these tasks with contrastive search using the top 5 candidates, except for RWKV for which we use sample search.

\begin{table}
\caption{Accuracy by model for predicting the next action  on the test set. (M = \gls{mn}, P = \gls{pid}, A = \gls{at})}
\label{tab:next_action_results}
\centering
\begin{tabular}{|l|r|r|r|r|}
\hline
Model & M & P & A & All \\
\hline
gpt2-25.3M & 0.274 & \textbf{0.130} & \textbf{0.527} & 0.016 \\
gpt2-46.5M & 0.266 & 0.114 & 0.357 & \textbf{0.018}\\
gpt2-89.0M & 0.225 & 0.075 & 0.372 & 0.007 \\
gpt2-131.6M & \textbf{0.293} & 0.122 & 0.489 & 0.014 \\
\hline
rwkv-65.7M & 0.136 & 0.050 & 0.200 & 0.005 \\
rwkv-127.2M & 0.011 & 0.023 & 0.031 & 0.001 \\
\hline
llama-58.1M & 0.177 & 0.085 & 0.337 & 0.012 \\
llama-112.0M & 0.156 & 0.039 & 0.407 & 0.012 \\
llama-219.8M & 0.174 & 0.038 & 0.381 & 0.001 \\
\hline
\end{tabular}
\end{table}

We evaluate next-action accuracy with a randomly sampled subset of sessions from the test set, which total around 82,685 \gls{al} events. For each \gls{al} event in the session beyond the first event, we use all preceding context to generate per-event predictions. We evaluate the accuracy of each feature, as well as the ability to generate correct predictions for all features, in \tableref{tab:next_action_results}.

We believe that ROUGE-1 is an appropriate metric since we wish to evaluate the recall performance of specific events. As with accuracy, we evaluate ROUGE-1 performance per-field and over the entire sequence, seen in \tableref{tab:rouge_scores}. For evaluation, we use 50\% of the input to generate predictions and use the remaining 50\% as a reference. We tested ROUGE-1 and ROUGE-L, and find that they are identical, possibly because the generated and reference sequences have no common order.

\begin{table}
\centering
\caption{ROUGE-1 scores with 5,000 randomly sampled test sessions. (M = \gls{mn}, P = \gls{pid}, A = \gls{at})}
\label{tab:rouge_scores}
\begin{tabular}{|l|r|r|r|r|}
\hline
Model & M & P & A & All \\
\hline
gpt2-25.3M & 0.168 & \textbf{0.159} & \textbf{0.463} & \textbf{0.264} \\
gpt2-46.5M & 0.249 & 0.088 & 0.319 & 0.219 \\
gpt2-89.0M & \textbf{0.254} & 0.072 & 0.333 & 0.220 \\
gpt2-131.6M & 0.213 & 0.095 & 0.337 & 0.215 \\
\hline
rwkv-65.7M & 0.042 & 0.051 & 0.254 & 0.116 \\
rwkv-127.2M & 0.003 & 0.075 & 0.009 & 0.029 \\
\hline
llama-58.1M & 0.045 & 0.064 & 0.169 & 0.093 \\
llama-112.0M & 0.059 & 0.019 & 0.448 & 0.175 \\
llama-219.8M & 0.165 & 0.042 & 0.357 & 0.188 \\
\hline
\end{tabular}
\end{table}

\subsection{Per-Feature Perplexity}

\begin{figure*}[htbp]
    \centering
    \includegraphics[width=0.7\textwidth]{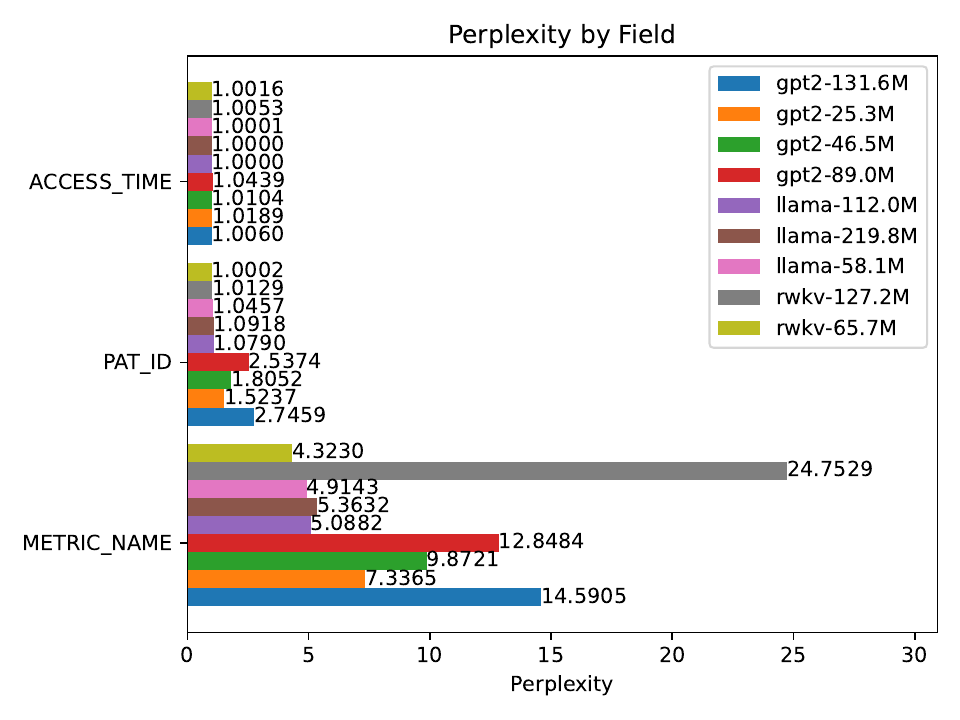}
    \caption{Perplexity of features by model.}
    \label{fig:ppl_features}
\end{figure*}

We evaluate the per-feature cross-entropy performance on the validation and test sets. The results from each field are shown in \figureref{fig:ppl_features}, and reported as perplexity for viewability. As expected, we find that \gls{mn} has the highest perplexity, and the LLaMA models appear to perform the best overall.

This suggests that an upper bound for the average number of possible successive \gls{mn} is approximately $4.3230$, which is remarkable given the $4,037$ distinct possibilities in the \gls{mn} 
vocabulary. 

\subsection{\Gls{ce} as a Measure of Workflow Complexity}

We include a few brief examples of the primary motivation behind the development of our \gls{al} \glspl{lm}, which is to measure the complexity in clinician workflow. We include an example of short sequence with moderate cross-entropy and pairs of repeating elements, a sequence with many repeating elements, and high entropy sequence in \appendixref{apd:examples}. It should be noted that the \gls{at} listed is the upper-bound of the quantized bin, and the \gls{pid} is its unique identifier within its parent shift. 

In the example shown in \tableref{tab:average_ent}, we have two repeating sets of actions, ``Notes viewed'' and ''Inpatient Patient Lists list loaded.'' In both cases, there is decreasing entropy in the latter element of the pairs. This is particularly true with the second instance of ``Notes viewed,'' which we believe is reflective of the fact that clinicians will often look through multiple clinical notes for a patient before they move onto the next action step in the care sequence.

In \tableref{tab:repeat_ent}, we present an example with not only a higher number of repeating elements, but also several system-generated automated events and unexpected events. For example, when ``Inpatient Patient Lists list loaded'' occurs it is always followed by a ``Inpatient system list accessed'' event, which is reflected in the low cross-entropy of the third row in \tableref{tab:repeat_ent}. We also see many repeated elements of ``Inpatient system list accessed'' and ``In Basket message viewed,'' and when their cycles are broken with ``In Basket folder loaded,'' ``Cosign clinical note,'' and ``View FastNote activity,'' we see a significant increase in the cross-entropy, indicating that they are essentially unexpected events.

The example in \tableref{tab:high_ent} is a case with high row cross-entropy. We believe that the entropy is high here because of the fact the session is taking place through Epic Haiku, which is a mobile client with its own subset of \gls{mn}, and so the actions performed have a much lower likelihood overall and more unstructured workflow compared to typical \gls{ehr} usage on a desktop workstation.






\section{Discussion}

We find that \glspl{al} can be modeled with considerable effectiveness using \glspl{lm} and that these \glspl{lm} can be used to measure the cross-entropy of \glspl{al}, particularly in relative comparisons. Although there are many limitations to the \gls{tlm} approach used, these models provide an initial insight into measuring the disorderedness in \glspl{al}, which may be applied to quantify the complexity of clinical workflow. While the generative performance is considerably weaker than in natural language, we believe this is due to the inherent structure of \glspl{al}, as they do not have all context of a clinical workflow, contain much more repetition, and have a more rigid structure than natural language. This may be part of the reason we find that higher parameter counts do not always translate into higher performance, as has generally been seen in other \glspl{lm} \citep{kaplan2020scaling}.



\subsection{Limitations} \label{sec:limitations}

A major design limitation of these models is the use of quantization, as the number of bins is effectively limited by the imbalanced number of examples, and requires significant manual tuning to determine the appropriate number of bins.

Another limitation of the \glspl{lm} presented here is that we train on a private dataset of \gls{icu} clinicians, which limits the transferability of the model to other types of clinicians with different specialties and roles, as some types of workflows may not appear at all in the training set. Future models should include a more diverse set of clinical specialties. Furthermore, we treat each session of \gls{ehr}-based work independently, under the assumption that clinicians who practice in similar specialty and settings have similar workflow, and potential individual differences in \gls{ehr} interaction styles are not considered as a result. 

Finally, the \glspl{al} used in this study captures all \gls{ehr}-based work during a fixed period of time from a set of clinicians who are assumed to perform similar workflow. However, there may be inherent heterogeneity of workflow being introduced from different clinician job levels or work locations, which were not explicitly addressed during this study.

\subsection{Future Work}
The \textit{validation} of cross-entropy as a measure of workflow complexity, either against existing proxies of characterizing workflow or through an expert manual review on a sample of action event sequences, will be a significant next step in this area.

Furthermore, training separate models after identifying different subgroups within the \glspl{al} may allow us to capture the non-routineness of clinicians' \gls{ehr}-based actions with better accuracy, which in turn can be better reflective description of the clinician workflow.

Although our original intent was to measure the cross-entropy of \glspl{al}, we expect these \glspl{tlm} can be effectively fine-tuned for a variety of different supervised tasks involving clinical workflow. Future work should also focus on improvements in the architectural design, especially with respect to the quantization of the \gls{pid} and \gls{at} \glspl{fv}; and on a clinician-centric training approach for a customized assessment of the effect of divergent behavioral patterns in evaluating work efficiencies at the individual clinician level.

\acks{This work was made possible with the support of grants 1R25LM014224-01 and R01LM013778 from the National Library of Medicine.}

\bibliography{gscholarent}

\appendix

\section{Examples}\label{apd:examples}

We highlight three different examples of a session of \gls{al}, as well as their per-row cross-entropies calculated using llama-112.0M using all preceding context. The first example in \tableref{tab:average_ent} is a randomly selected example with average cross-entropy near 1 and several repeating elements. The next example in \tableref{tab:repeat_ent} is a session with many repeating and automated events. Finally, we include a session with high cross-entropy in \tableref{tab:high_ent}.

\begin{table*}
    \centering
    \begin{tabular}{llll}
    \toprule
    \gls{mn} & \gls{pid} & \gls{at} & Row Entropy \\
    \midrule
Inpatient Patient Lists list loaded & -1 & $\leq$ 1 & - \\
Results Review accessed & 2 & 3.936 & 1.766 \\
Notes viewed & 2 & 15.492 & 1.247 \\
Notes viewed & 2 & 3.936 & 0.266 \\
Notes viewed & 2 & 3.936 & 0.222 \\
Results Review exited & 2 & 15.492 & 1.249 \\
Results Review exited & 1 & $\leq$ 1 & 1.249 \\
    \bottomrule
    \end{tabular}
    \caption{An example of a session with cross-entropy near 1 and several repeating elements.}
    \label{tab:average_ent}
\end{table*}

\begin{table*}
    \centering
\begin{tabular}{llll}
\toprule
    \gls{mn} & \gls{pid} & \gls{at} & Row Entropy \\
\midrule
Inpatient Patient Lists list loaded & -1 & $\leq$ 1 & - \\
Inpatient system list accessed & -1 & $\leq$ 1 & 0.111 \\
Inpatient system list accessed & -1 & $\leq$ 1 & 0.289 \\
Inpatient system list accessed & -1 & $\leq$ 1 & 0.179 \\
In Basket folder loaded & -1 & 15.492 & 1.332 \\
In Basket folder loaded & -1 & $\leq$ 1 & 0.886 \\
In Basket message viewed & 3 & 3.936 & 0.068 \\
View FastNote activity & 3 & 3.936 & 1.844 \\
Cosign clinical note with attestation & 3 & 15.492 & 1.544 \\
Problem List accessed & 3 & $\leq$ 1 & 1.353 \\
In Basket message viewed & 3 & 3.936 & 0.782 \\
In Basket message viewed & 3 & 3.936 & 0.269 \\
In Basket message viewed & 15 & 3.936 & 0.121 \\
Cosign clinical note & 15 & 3.936 & 0.785 \\
In Basket message viewed & 15 & $\leq$ 1 & 0.052 \\
In Basket message viewed & 0 & $\leq$ 1 & 0.220 \\
Cosign clinical note & 0 & 3.936 & 0.512 \\
In Basket message viewed & 0 & $\leq$ 1 & 0.051 \\
In Basket message viewed & 3 & $\leq$ 1 & 0.535 \\
\bottomrule
\end{tabular}
    \caption{An example of a session with highly cyclical events, as well as interrupting events.}
    \label{tab:repeat_ent}
\end{table*}

\begin{table*}
    \centering
    \begin{tabular}{llll}
    \toprule
    \gls{mn} & \gls{pid} & \gls{at} & Row Entropy \\
    \midrule
Haiku login & -1 & $\leq$ 1 & - \\
Inpatient system list accessed & -1 & 3.936 & 1.232 \\
Report viewed & 116 & 3.936 & 4.703 \\
Results List viewed & 116 & 3.936 & 5.167 \\
Report viewed & 116 & 3.936 & 5.041 \\
Report viewed & 116 & 3.936 & 5.377 \\
Results List viewed & 116 & 3.936 & 4.921 \\
Report viewed & 116 & 3.936 & 4.033 \\
Report viewed & 116 & 3.936 & 5.724 \\
Results table viewed & 116 & 3.936 & 5.831 \\
Results table viewed & 116 & 3.936 & 5.121 \\
Haiku login & -1 & 15.492 & 2.508 \\
    \bottomrule
    \end{tabular}
    \caption{An example of a session with some high cross-entropy elements.}
    \label{tab:high_ent}
\end{table*}

\section{Model Card}\label{apd:first}

\tableref{tab:model_card} contains a model card \citep{mitchell2019model, touvron2023llama, anil2023palm} with information for users who wish to utilize one of our models.

\begin{table*}[h]
    \centering
    \begin{tabular}{|l|p{4in}|}
        \hline
        \multicolumn{2}{|c|}{\textbf{Model Details}}\\
        \hline
         Model Developers & Washington University in St. Louis \\
         \hline
         Variations & GPT-2 (25.3M, 46.5M, 89.0M, 131.6M), RWKV (65.7M, 113.5M), LLaMA (58.1M, 112.0M, 219.8M)  \\
         \hline
         Input & Epic \acrlong{al} data, up to 1024 tokens (341 complete rows) \\
         \hline
         Output & Tokens representing generated Epic \acrlong{al} events in the \gls{mn}, \gls{pid}, and \gls{at} columns.\\
         \hline
         Model Architecture & {
            All GPT-2: 6 attention heads;
            gpt2-25.3M: 3 layers;
            gpt2-46.5M: 6 layers;
            gpt2-89.0M: 12 layers;
            gpt2-131.6M: 18 layers;
            All RWKV: 512 attention hidden size, 2048 intermediate size; 
            rwkv-65.7M: 18 layers;
            rwkv-127.2M: 36 layers;
            All LLaMA: 32 attention heads, 512 hidden size;
            llama-58.1M: 3 layers; llama-112.0M: 6 layers; llama-219.8M: 12 layers
         }\\
         \hline
         Model Dates & These models were trained in September and October 2023. \\
         \hline
         Status & This model was trained with a private dataset of \gls{icu} clinician audit logs. We do not anticipate releasing updates to our model, however further studies involving different clinician types may result in new models. \\
         \hline
         License & Software: Apache 2.0, Model Weights: Apache 2.0 \\
         \hline
         Where to send comments & Up-to-date instructions for leaving comments can will be in the README of our GitHub repo at \url{https://github.com/bcwarner/audit-log-lm} \\
         \hline
        \multicolumn{2}{|c|}{\textbf{Intended Use}}\\ 
        \hline
        Intended Use Cases & Our models are intended for research use with Epic \acrlongpl{al}, either for generative modeling or for cross-entropy measurement. \\
        \hline
        Out-of-Scope Usage & Because of the experimental nature of these models, we do not recommend the usage of our models in a non-research context. \\
        \hline
        \multicolumn{2}{|c|}{\textbf{Hardware and Software}}\\
        \hline
        Training & We used Hugging Face \texttt{transformers} \citep{wolf2019huggingface} for the model architectures atop PyTorch \citep{paszke2019pytorch} and PyTorch Lightning, and we use Sophia to optimize \citep{liu2023sophia}. We trained on a mixture of NVIDIA A40s, 80GB A100s, and GeForce RTX 2080s in a research compute cluster maintained at Washington University in St. Louis. \\
        \hline
        \multicolumn{2}{|c|}{\textbf{Training Data}}\\
        \hline
        Overview & These models were trained on a set of \gls{al} obtained from \gls{icu} physicians performing work in the BJC HealthCare/Washington University system as a part of the ASPIRE study (IRB\# 202009032, NLM grant 5R01LM013778-03). The original training data is \gls{phi} and cannot be released. \\
        \hline
        Data Freshness & The training data was taken over a period of March 2019 to February 2020 (pre-COVID).
        \\
        \hline
        \multicolumn{2}{|c|}{\textbf{Evaluation Results}}\\
        \hline
        \multicolumn{2}{|l|}{See \sectionref{sec:results} for details on the performance of each of our models.}\\
        \hline
        \multicolumn{2}{|c|}{\textbf{Ethical Considerations and Limitations}}\\
        \hline
        \multicolumn{2}{|p{5.5in}|}{Because of their experimental nature, we do not recommend the usage of these models outside of a research context, due to the limitations discussed in \sectionref{sec:limitations}. In addition, no generated data will contain \gls{phi}, as the the two \gls{phi} columns (\gls{pid} and \gls{at}) have been quantized, and clinician identifiers are not used as a feature.}
        \\
        \hline
    \end{tabular}
    \caption{Model card for the models discussed.}
    \label{tab:model_card}
\end{table*}

\end{document}